\documentclass[journal=jacsat,manuscript=article]{achemso}

\usepackage{natbib} 
\usepackage[version=3]{mhchem}
\usepackage{geometry}
\usepackage{balance}
\usepackage{mathptmx}
\usepackage{sectsty}
\usepackage{graphicx} 
\usepackage{lastpage}
\usepackage[format=plain,justification=justified,singlelinecheck=false,font={stretch=1.125,small,sf},labelfont=bf,labelsep=space]{caption}
\usepackage{float}
\usepackage{fancyhdr}
\usepackage{fnpos}
\usepackage{siunitx}
\usepackage[english]{babel}
\addto{\captionsenglish}{%
  
}
\usepackage{array}
\usepackage{droidsans}
\usepackage{charter}
\usepackage[T1]{fontenc}
\usepackage[usenames,dvipsnames]{xcolor}
\usepackage{setspace}
\usepackage[compact]{titlesec}
\usepackage{hyperref}
\usepackage{epstopdf}%This line makes .eps figures into .pdf - please comment out if not required.

\author{Reza Reihanisaransari}\affiliation[University of Houston]
{Department of Electrical and Computer Engineering, University of Houston, Houston, TX}
\altaffiliation{Both authors contributed equally to this work}
\author{Chalapathi Charan Gajjela}\affiliation[University of Houston]
{Department of Electrical and Computer Engineering, University of Houston, Houston, TX}
\altaffiliation{Both authors contributed equally to this work}
\author{Xinyu Wu}\affiliation[University of Houston]
{Department of Electrical and Computer Engineering, University of Houston, Houston, TX}
\author{Ragib Ishrak}\affiliation[University of Houston]
{Department of Electrical and Computer Engineering, University of Houston, Houston, TX}

\author{Sara Corvigno}\affiliation[MD Anderson]
{The University of Texas MD Anderson Cancer Center, Houston, TX 77030, USA}

\author{Yanping Zhong}\affiliation[MD Anderson]
{The University of Texas MD Anderson Cancer Center, Houston, TX 77030, USA}

\author{Jinsong Liui}\affiliation[MD Anderson]
{The University of Texas MD Anderson Cancer Center, Houston, TX 77030, USA}

\author{Anil K. Sood}\affiliation[MD Anderson]
{The University of Texas MD Anderson Cancer Center, Houston, TX 77030, USA}

\author{David Mayerich}\affiliation[University of Houston]
{Department of Electrical and Computer Engineering, University of Houston, Houston, TX}

\author{Sebastian Berisha}\affiliation[University of Houston]
{Milwaukee School of Engineering, Milwaukee, WI, USA}

\author{Rohith Reddy}\affiliation[University of Houston]
{Department of Electrical and Computer Engineering, University of Houston, Houston, TX}
\altaffiliation{Address, 4226 Martin Luther King Boulevard,  N308 Engineering Building 1, Houston TX, 77584, USA}
\email{rkreddy@uh.edu}

\title[An \textsf{achemso} demo]
  {Rapid hyperspectral photothermal mid-infrared spectroscopic imaging from sparse data for gynecologic cancer tissue subtyping}

\abbreviations{IR,NMR,UV}

\begin{document}


\begin{abstract}
Ovarian cancer detection has traditionally relied on a multi-step process that includes biopsy, tissue staining, and morphological analysis by experienced pathologists. While widely practiced, this conventional approach suffers from several drawbacks: it is qualitative, time-intensive, and heavily dependent on the quality of staining. Mid-infrared (MIR) hyperspectral photothermal imaging is a label-free, biochemically quantitative technology that, when combined with machine learning algorithms, can eliminate the need for staining and provide quantitative results comparable to traditional histology. However, this technology is slow. This work presents a novel approach to MIR photothermal imaging that enhances its speed by an order of magnitude. Our method significantly accelerates data collection by capturing a combination of high-resolution and interleaved, lower-resolution infrared band images and applying computational techniques for data interpolation. We effectively minimize data collection requirements by leveraging sparse data acquisition and employing curvelet-based reconstruction algorithms. This approach enhances imaging speed without compromising image quality and ensures robust tissue segmentation. This method resolves the longstanding trade-off between imaging resolution and data collection speed, enabling the reconstruction of high-quality, high-resolution images from undersampled datasets and achieving a 10X improvement in data acquisition time. We assessed the performance of our sparse imaging methodology using a variety of quantitative metrics, including mean squared error (MSE), structural similarity index (SSIM), and tissue subtype classification accuracies, employing both random forest and convolutional neural network (CNN) models, accompanied by Receiver Operating Characteristic (ROC) curves. Our statistically robust analysis, based on data from 100 ovarian cancer patient samples and over 65 million data points, demonstrates the method's capability to produce superior image quality and accurately distinguish between different gynecological tissue types with segmentation accuracy exceeding 95\%. Our work demonstrates the feasibility of integrating rapid MIR hyperspectral photothermal imaging with machine learning in enhancing ovarian cancer tissue characterization, paving the way for quantitative, label-free, automated histopathology. It represents a significant leap forward from traditional histopathological methods, offering profound implications for cancer diagnostics and treatment decision-making.
\end{abstract}

\section{Introduction}
Mid-infrared spectroscopic imaging (MIRSI) is a class of quantitative, label-free, non-destructive techniques for acquiring spatially resolved chemical information from a sample. Its utility extends across various fields, such as disease diagnosis, offering an alternative to histopathology \cite{querido2021applications,hirschmugl2012fourier, mankar2022polarization, reddy2010accurate, walsh2012label, pahlow2018application, baker2014using, fernandez2005infrared, gosling2023microcalcification}, as well as material science \cite{xu_ftir_2019, prati2010new, qin2020spontaneous}, environmental and toxicological chemistry \cite{trevisan2012extracting, simonescu2012application}, and forensics \cite{ewing2017infrared, ricci2006enhancing}. Fourier transform infrared (FT-IR) spectroscopic imaging is the best-known MIRSI technology and has been the {\it de facto} standard for spatially resolved molecular fingerprinting of organic molecules \cite{pahlow2018application, baker2014using,fernandez2005infrared, pounder2016development}. FT-IR measurements typically cover 800-4000 $cm^{-1}$ MIR wavenumbers. However, the acquisition process is notably slow, as not every wavenumber offers distinct chemical information. Additionally, the resolution of FT-IR is constrained by diffraction limits \cite{reddy2013high}. For effective analysis, samples must be thin (around $5 \ \mu m$) and dehydrated due to the substantial challenges posed by water absorption. Previous research has demonstrated that only a certain subset of wavenumbers contain features necessary for deciphering the chemical composition of samples \cite{zohdi_importance_2015, lotfollahi2022adaptive, deutsch2015compositional}. The adoption of Quantum Cascade Laser (QCL)-based Discrete Frequency IR (DFIR) imaging mitigates some of the limitations of FT-IR imaging by facilitating data acquisition at fewer wavenumbers, specifically those with chemically significant features \cite{kole2012discrete, yeh2019multicolor, yeh2023infrared, yeh2015fast}. The tunability and wavenumber selectivity offered by QCL sources enable DFIR instruments to acquire data at specific wavenumbers tailored to the application, thereby enhancing the speed of data acquisition. Despite these advancements, both DFIR and FT-IR are subject to a diffraction-limited spatial resolution of \SI{5.5}{\micro\meter}. 

\begin{figure}[hbt]
    \centering
  \includegraphics[width=0.8\linewidth]{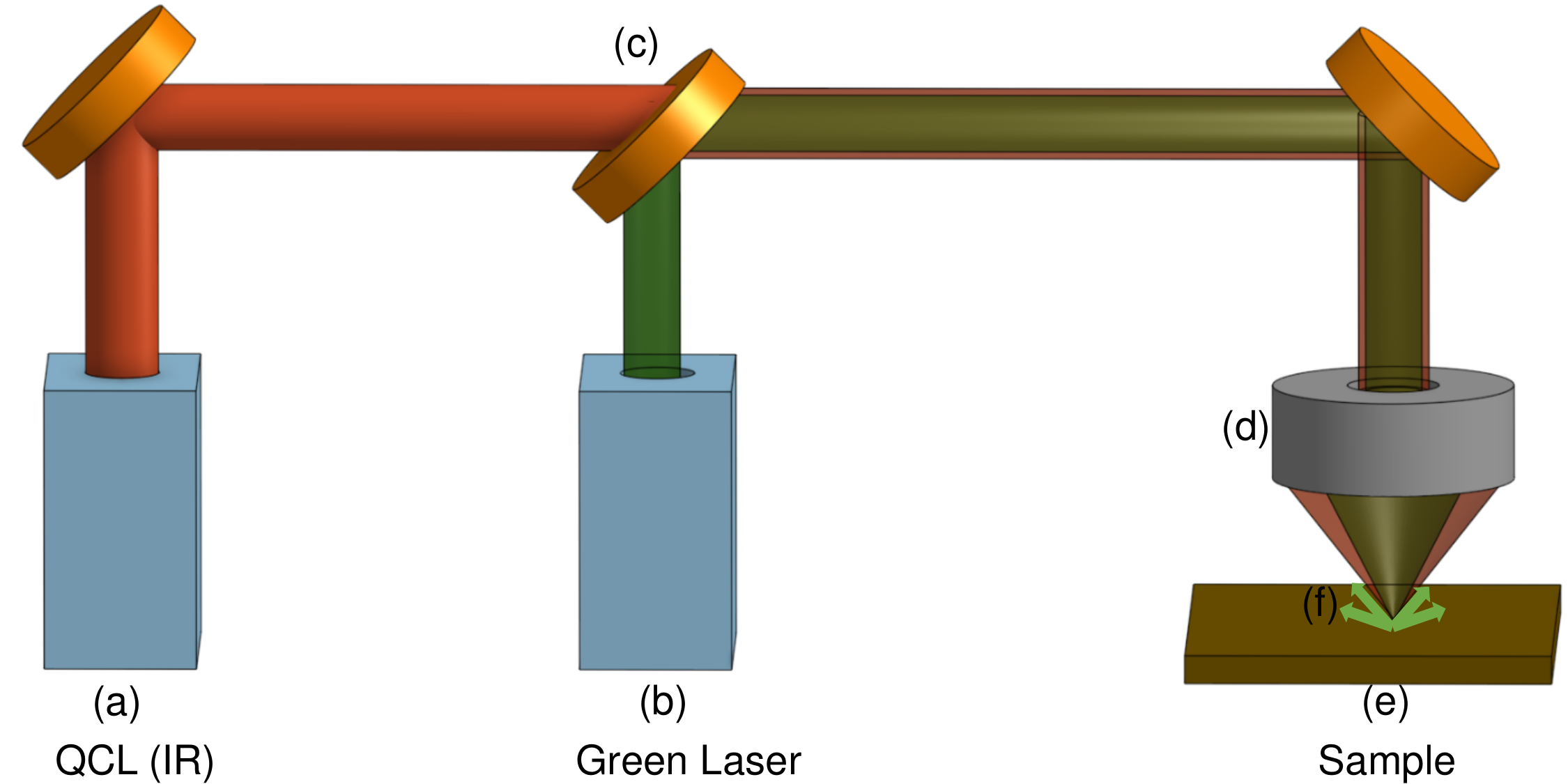}\caption{
    Schematic illustration of the O-PTIR optical configuration showing both the IR and green (532 nm) laser paths. Pulsed QCL at point (a) causes a photothermal expansion in the sample. A Continuous Wave (CW) green laser, indicated by (b), is collinearly directed onto the sample to serve as a probe beam. A dichroic mirror (c) merges the green and QCL beams, focusing them onto the sample (e) through a reflective Cassegrain objective (d). The resulting modulation in the intensity of the green light (f), scattered back from the sample, facilitates the measurement of its IR absorbance.}
    \label{fig:schematic}
\end{figure}

\begin{figure}[!hbt]
    \centering
  \includegraphics[width=0.8\linewidth]{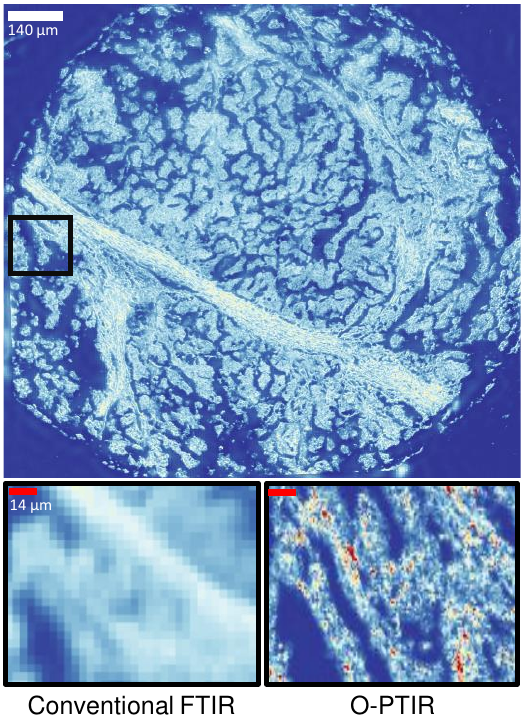}
    \caption{Comparison of High-Definition FT-IR and O-PTIR images of a cancerous Core. This figure illustrates the significant advantage of O-PTIR over FT-IR, showcasing its ability to overcome the diffraction limit, which results in enhanced spatial resolution. The improved image quality of O-PTIR is evident.}
    \label{fig:ptir_ftir}
\end{figure}

The introduction of Optical Photothermal Infrared (O-PTIR) imaging \cite{zhang2016depth, bialkowski1996photothermal, bai2019ultrafast, xia2022mid, bai2021bond} overcomes resolution limitations by providing a \SI{0.5}{\micro\meter} spatial resolution and delivers information 100 times more detailed than that provided by FT-IR. O-PTIR imaging overcomes the IR diffraction limit using a pump and probe mechanism. The IR-induced photothermal effect alters the sample's optical properties, leading to changes in visible light intensity, which is proportional to the IR absorption of infrared radiation. Detection is achieved through a coaxial and confocal visible (532 nm) light probe illustrated in Figure\ref{fig:schematic}. Figure \ref{fig:ptir_ftir} compares the image quality of O-PTIR, FT-IR on the same cancer tissue. Cropped region in this figure is $140\times140 \ \mu m$. At $5\times5 \ \mu m$ resolution we have $28\times28$ pixels which is the limit for FTIR. On the other hand, same region for O-PTIR at $0.5\times0.5 \ \mu m$ resolution has $280\times280$ pixels. The improved spatial resolution of O-PTIR relative to FT-IR is evident. However, this superior resolution results in extended data acquisition times.

To successfully apply O-PTIR imaging in clinical settings, faster acquisition speeds are required, which are currently not achievable. One possible solution is to exploit the high data redundancy in MIRSI through sparse data acquisition and subsequent reconstruction, significantly shortening the data acquisition time by orders of magnitude. Research across various modalities has demonstrated the feasibility of reconstructing data using diverse sparse sampling algorithms\cite{mankar2021multi}. This paper proposes using non-uniform rectangular sampling for data acquisition, along with curvelet-based reconstruction, to dramatically improve data acquisition speed.

The following discussion highlights acquisition speed challenges with current commercial O-PTIR systems and also suggests potential solutions. O-PTIR imaging employs raster scanning, with imaging time directly proportional to the image's height (Y dimension). Table \ref{tab: X-Y sampling vs. imaging time} demonstrates that utilizing rectangular data sampling can reduce imaging time. Increasing the spacing for data sampling along the Y-dimension means less data is collected compared to uniformly sampled data at high resolution across both X and Y dimensions. The third column of the table presents the percentage of data acquired relative to the original high-resolution images. This non-uniform sampling method coupled to reconstruction algorithms \cite{candes2008introduction} can effectively reduce acquisition time by leveraging the spatial and spectral sparsity inherent in MIRSI data \cite{deutsch2015compositional}.

Validating our approach with multiple methodologies is essential for ensuring the robustness and generalizability of our methods. Therefore, we propose three independent metrics for assessing reconstruction accuracy: mean square error (MSE), structural similarity index measure (SSIM), and classification accuracy. MSE quantifies the average discrepancy between the reconstructed images and the original, ground-truth images. In contrast, SSIM evaluates the visual similarity and the presence of artifacts in the reconstructed images. The application of machine learning algorithms is pivotal in various domains, ranging from electronics\cite{reihanisaransari_reliability_2022} to cancer diagnosis\cite{sabzalian_new_2023}. Given that one primary objective of our reconstruction is to enhance the segmentation accuracy of different cell types, we have employed machine learning algorithms and assessed their classification accuracy as an additional metric to ensure optimal reconstruction performance. We obtain data at multiple pixel spacing, measure reconstruction accuracies using the aforementioned metrics, and optimize our algorithms to achieve reliable performance. This reconstruction approach represents a novel and promising method capable of accelerating the acquisition of high-resolution spectroscopic data tenfold, thereby unlocking the full capabilities of the O-PTIR system. 

\begin{table}[tbh]
\centering
\begingroup
\setlength{\tabcolsep}{10pt} % Default value: 6pt
\renewcommand{\arraystretch}{1.5} % Default value: 1
\caption{Sample X-Y versus time for single band imaging (minutes needed for imaging $1500 \times 1500 \ \mu m$ region). As the pixel spacing increases, the acquisition time decreases, as shown in the second column. The higher pixel spacing results in some data being missing, as a tradeoff.}

\begin{tabular}{ccc}
\hline

X-Y spacing  &  Imaging time & Data fraction \\
($\mu m$ $\times$ $\mu m$) & (minutes) & (\%) \\
\hline
\SI{0.5}{\micro\meter} $\times$ \SI{0.5}{\micro\meter}   &     90      & 100\%                                                                                                                                                 \\
\SI{0.5}{\micro\meter}$\times$\SI{1}{\micro\meter}     &     45      & 50\%                                                                                                                                                     \\
\SI{0.5}{\micro\meter}$\times$\SI{2}{\micro\meter}     &     23     &  25\%                                                                                                                                                       \\
\SI{0.5}{\micro\meter}$\times$\SI{3}{\micro\meter}     &     15      & 15\%                                                                                                                                                  \\
\SI{0.5}{\micro\meter}$\times$\SI{5}{\micro\meter}     &    9      & 10 \%                                                                                                                                                 \\
\SI{0.5}{\micro\meter}$\times$\SI{10}{\micro\meter}    &    4.5     & 5  \%                                                                                                                                                 \\ 
\SI{0.5}{\micro\meter}$\times$\SI{20}{\micro\meter}    &    2.5      &2.5 \%                                                                                  \\ \hline
\end{tabular}
\label{tab: X-Y sampling vs. imaging time}
\endgroup
\end{table}

\section{Materials and Methods}
An ovarian biopsy tissue microarray (TMA) was obtained from Biomax US (BC11115c) and imaged using a commercial O-PTIR system (Mirage, Photothermal Spec.). The TMA consists of paraffin-embedded cores mounted on a 1 mm thickness CaF$_2$ substrate. These cores are from separate patients with cases of normal, hyperplastic, dysplastic, and malignant tumors. The patient cohort was composed of women aged 29 to 69; ovarian tumor stages varied between stage I to stage IIIC; histological subtypes include clear cell carcinoma, high-grade serous carcinoma, and Mucinous adenocarcinoma.  The deparaffinization was done following the protocol along the lines described in Baker {\it et al.}~\cite{baker2014using} before undergoing O-PTIR imaging. The paraffin-embedded samples were deparaffinized by washing the sample in 100\% xylene twice for 5 minutes each and then with 100\% ethanol thrice. The corresponding adjacent histological section was stained with H\&E and examined by an expert pathologist. Cell subtypes were identified across disease stages. We trained a random forest (RF) classifier, and a CNN model by using the 45 cores on the left half of TMA for training and testing on the remaining 55 cores on the right half of TMA, ensuring that we have an appropriate amount of pixels for each class in training and testing.

%\indent

\subsection{Data acquisition}
\begin{figure}[!hbt]
  \includegraphics[width=\linewidth]{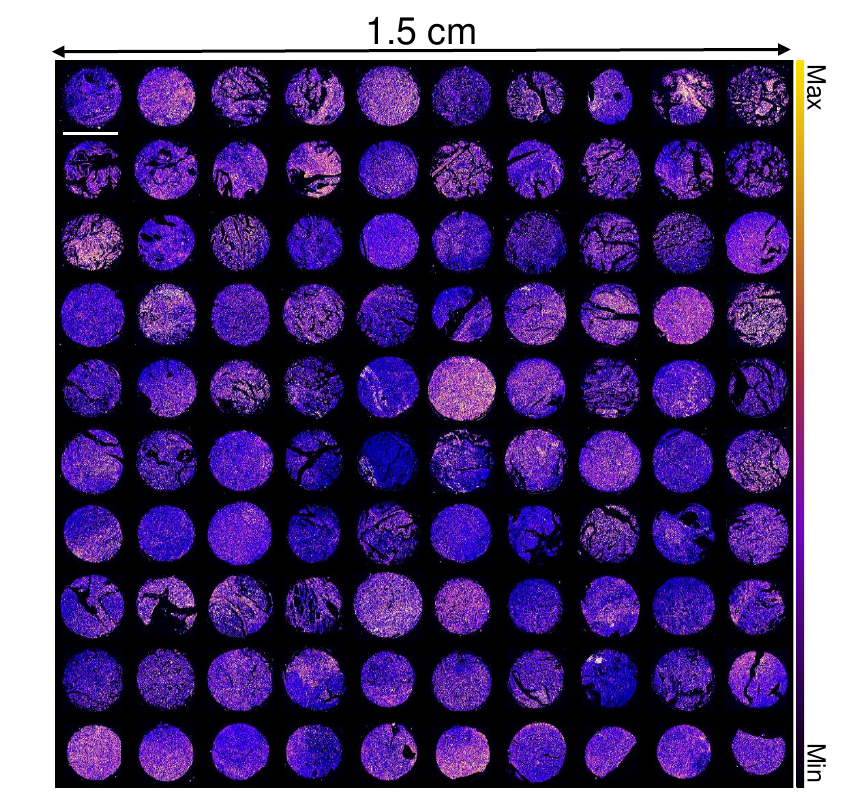}
    \caption{
    Microarray of ovarian cancer cores imaged by O-PTIR at the \SI{1660}{\per\centi\meter} band. The data encompasses samples from 100 ovarian cancer patients. Variations in tissue biochemistry are highlighted by the color differences, demonstrating the rich biochemical information at the \SI{1660}{\per\centi\meter} band, chosen for high-resolution reconstruction due to its significance in the fingerprint region. Scale bar: 1.5 mm.}
    \label{fig:ir-tma}
\end{figure}

The O-PTIR dataset was acquired using a Photothermal mIRage microscope with a silicon photodiode, a pixel size of \SI{0.5}{\micro\meter}$\times$\SI{0.5}{\micro\meter} and a $0.65$ numerical aperture. A Quantum Cascade Laser (QCL) source sweeps through the range of \SI{902} {\per\centi\meter} to \SI{1898} {\per\centi\meter}. Each core was imaged at 28 selected wavenumbers (\SI{908}{\per\centi\meter}, \SI{974}{\per\centi\meter}, \SI{984}{\per\centi\meter}, \SI{1036}{\per\centi\meter}, \SI{1070}{\per\centi\meter}, \SI{1102}{\per\centi\meter}, \SI{1136}{\per\centi\meter}, \SI{1178}{\per\centi\meter}, \SI{1238}{\per\centi\meter}, \SI{1280}{\per\centi\meter}, \SI{1300}{\per\centi\meter}, \SI{1325}{\per\centi\meter}, \SI{1358}{\per\centi\meter}, \SI{1396}{\per\centi\meter}, \SI{1420}{\per\centi\meter}, \SI{1456}{\per\centi\meter}, \SI{1482}{\per\centi\meter}, \SI{1500}{\per\centi\meter}, \SI{1536}{\per\centi\meter}, \SI{1556}{\per\centi\meter}, \SI{1596}{\per\centi\meter}, \SI{1610}{\per\centi\meter}, \SI{1662}{\per\centi\meter}, \SI{1668}{\per\centi\meter}, \SI{1682}{\per\centi\meter}, \SI{1746}{\per\centi\meter}, and \SI{1786}{\per\centi\meter} ). Amide I band (\SI{1660}{\per\centi\meter}) was collected at high resolution (\SI{0.5}{\micro\meter}$\times$\SI{0.5}{\micro\meter} pixel size) and the remaining 27 bands were collected at (\SI{0.5}{\micro\meter}$\times$\SI{5}{\micro\meter} pixel size.) for the entire TMA. Image of the TMA acquired at the Amide I band is shown in Figure \ref{fig:ir-tma}. Background spectra are collected with 8 co-additions and used to normalize the raw data to calculate the IR absorbance at each band. We also collected these bands for 4 random cores at different spacing in Y-axis (\SI{0.5}{\micro\meter}$\times$\SI{0.5}{\micro\meter}, \SI{0.5}{\micro\meter}$\times$\SI{1}{\micro\meter},
\SI{0.5}{\micro\meter}$\times$\SI{2}{\micro\meter},
\SI{0.5}{\micro\meter}$\times$\SI{3}{\micro\meter},
\SI{0.5}{\micro\meter}$\times$\SI{5}{\micro\meter},
\SI{0.5}{\micro\meter}$\times$\SI{10}{\micro\meter},
\SI{0.5}{\micro\meter}$\times$\SI{20}{\micro\meter}) in order to calculate MSE and SSIM to identify the optimal pixel spacing for effective image reconstruction.

The adjacent H\&E stained TMA was imaged with a Nikon inverted optical microscope with a 10X, 0.4NA objective in the brightfield mode, and has diffraction-limited spatial resolution in the visible range (\SI{0.4}{\micro\meter} - \SI{0.7}{\micro\meter}). 
%\indent 

\subsection{Sparse Image Reconstruction}
\begin{figure}[!hbt]
    \centering
  \includegraphics[width=\linewidth]{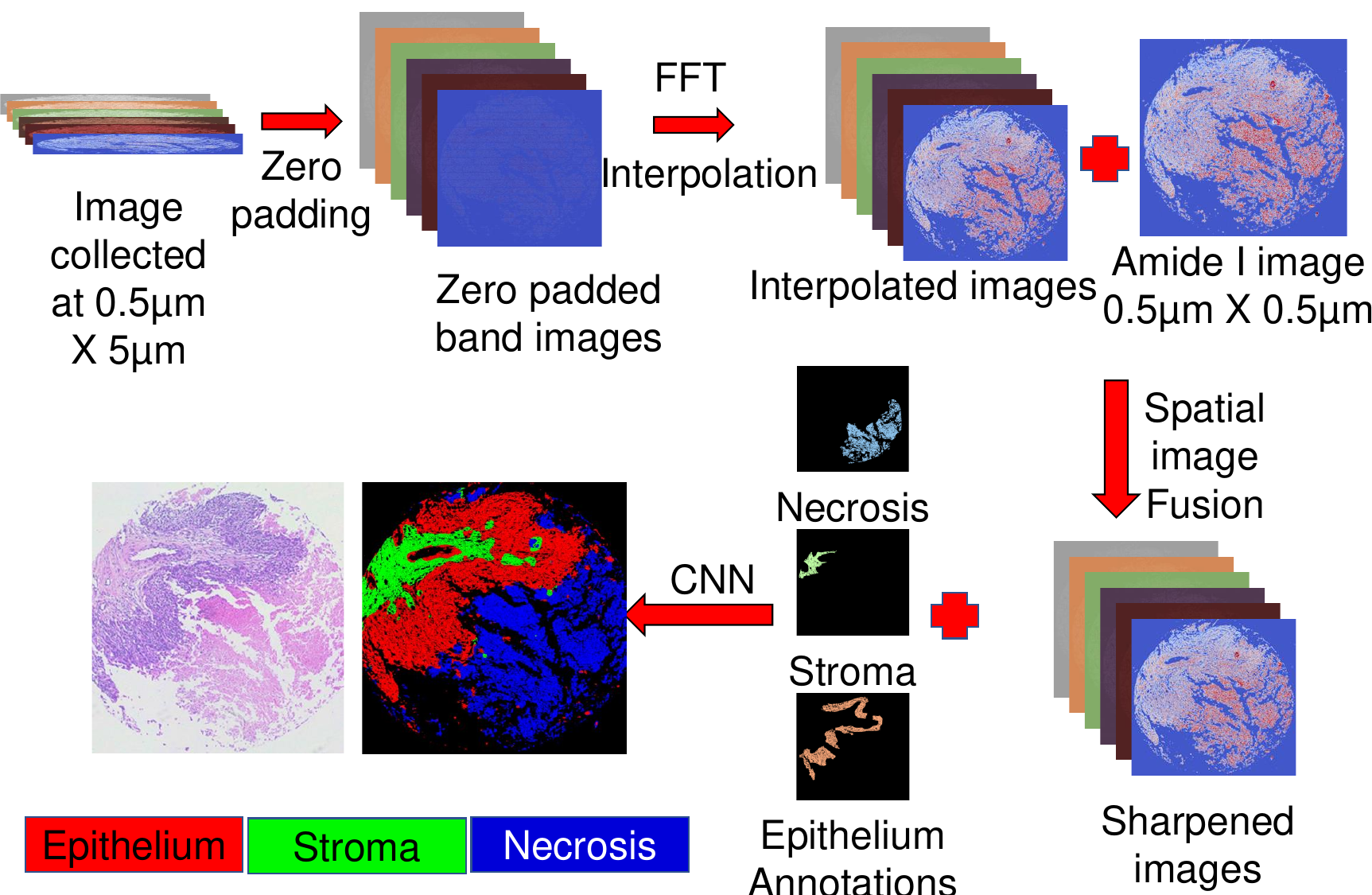}
    \caption{Schematic for the data reconstruction algorithm used to enhance data acquisition speed. The figure illustrates how rectangular pixel-spaced data (0.5X5) acquired from the O-PTIR system is used to reconstruct 27 high-resolution, diffraction-limited band images. This method increases the data acquisition speed by 10X, yielding high-resolution images that offer more detailed information for improved segmentation of different cell types. The algorithm fuses spatial features from a high-resolution Amide I image with the linearly interpolated rectangular image via curvelet transform. This fusion preserves the biochemical information of each band image while accurately translating the spatial features of biological samples.}
    \label{fig:schematic_flash_v1}
\end{figure}

We imaged tissue cores using sparse sampling along the y-axis to reduce O-PTIR imaging time, resulting in rectangular hyperspectral images. Using the curvelet transform, we reconstructed images to match the best resolution afforded by O-PTIR. These images were resized, registered, and then enhanced using an unsupervised curvelet transform, as illustrated in Figure \ref{fig:schematic_flash_v1}. The images were acquired with a \SI{0.5}{\micro\meter} spacing along the x-axis and variable spacing along the y-axis, ranging from \SI{0.5}{\micro\meter} to \SI{20}{\micro\meter}.

\subsubsection{Interpolation}

\begin{figure}[!hbt]
    \centering
  \includegraphics[width=\linewidth]{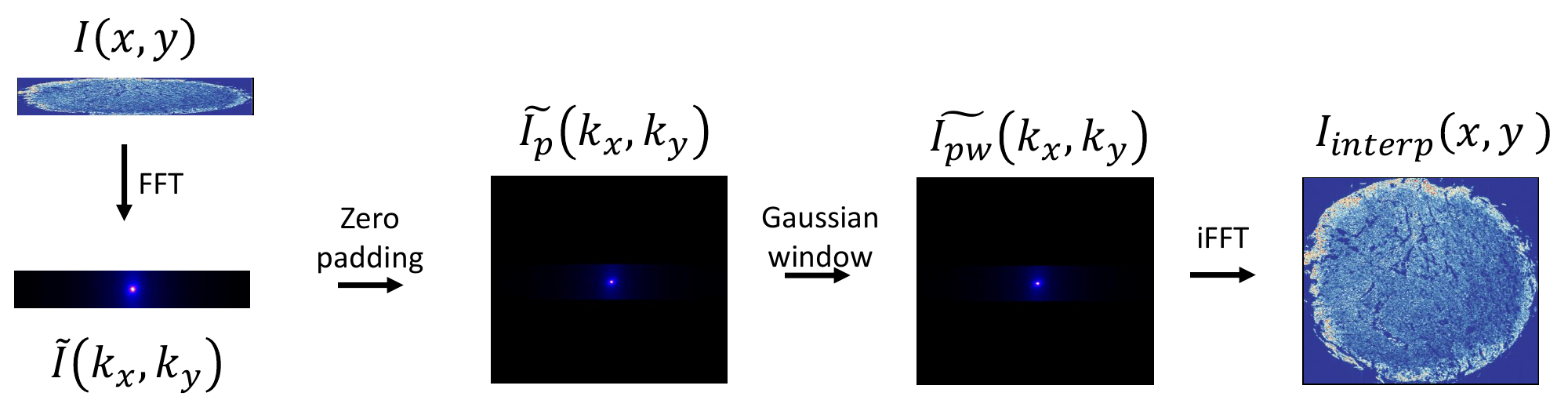}
    \caption{
 Method for interpolating low-resolution band images. We begin by performing a Fourier transform, followed by padding zeros along the Y-axis, and then applying a Gaussian filter to isolate lower frequencies. The interpolated image is obtained by taking the inverse Fourier transform.}
    \label{fig:schem_inter}
\end{figure}

To reconstruct images, we initially rescale the raw rectangular images along the y-dimension to match a pixel size of \SI{0.5}{\micro\meter}$\times$\SI{0.5}{\micro\meter}. This process involves computing the Fourier transform of each low-resolution (\SI{0.5}{\micro\meter}$\times$\SI{5}{\micro\meter}) band, then centering the lower frequencies in the Fourier domain. We utilize the high-resolution (\SI{0.5}{\micro\meter}$\times$\SI{0.5}{\micro\meter}) Amide I band (1660 cm$^{-1}$) as a reference for determining the interpolated image's dimensions. Subsequently, we zero-pad the low-resolution image along the y-axis to align with the high-resolution image's size. After padding, we apply a Gaussian window to smooth the image. The interpolated image is finally obtained by performing the inverse Fourier transform. Please see Figure \ref{fig:schem_inter} for a visual overview of the interpolation process.

\subsubsection{Curvelet Transform}
\begin{figure}[!hbt]
    \centering
  \includegraphics[width=\linewidth]{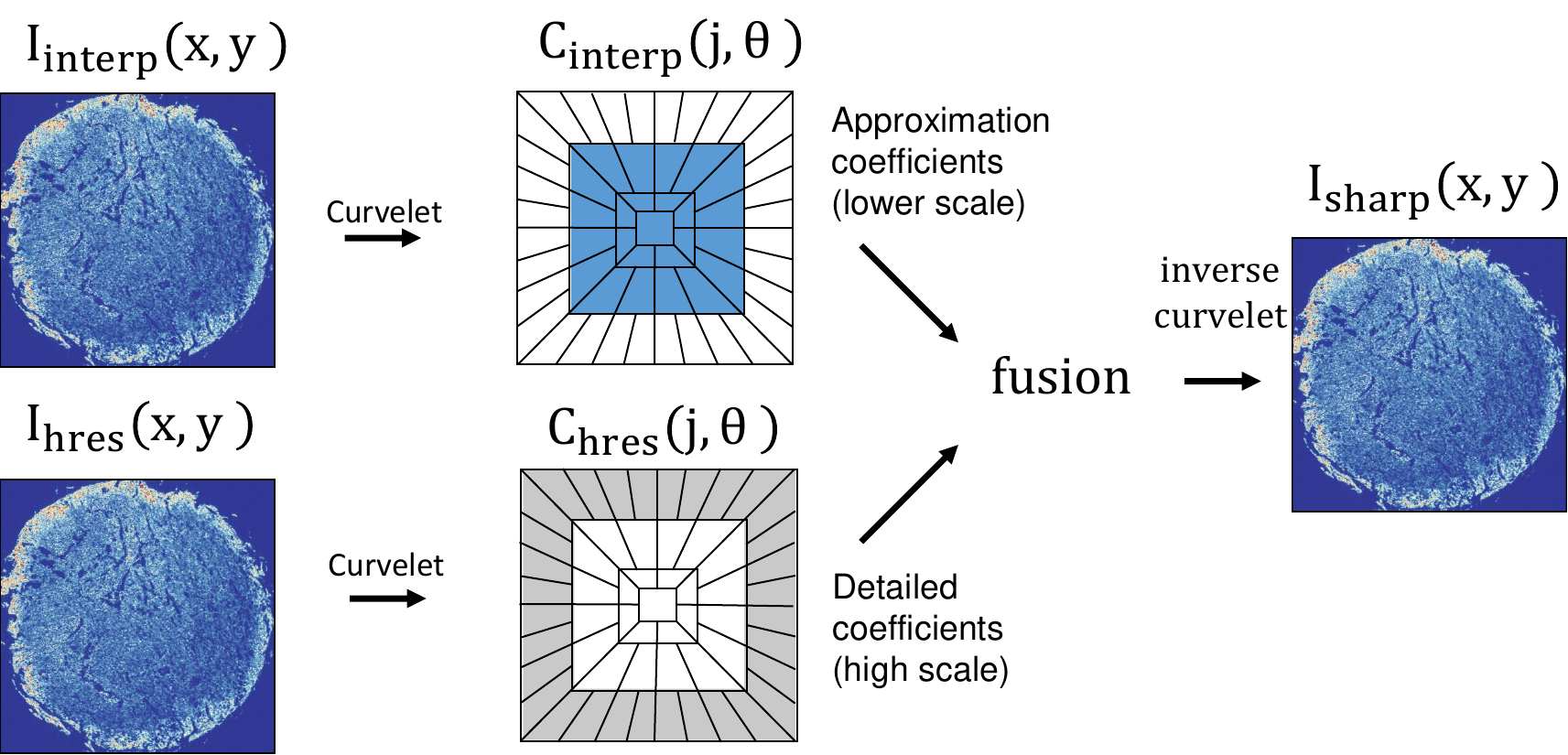}
    \caption{Data fusion through curvelet transform involves taking the curvelet transform of both low-resolution band images and the high-resolution Amide I image. We obtain the high-frequency coefficients from the high-resolution image to achieve sharp edges, while low-frequency coefficients are obtained from the low-resolution band image. By combining these two and taking the inverse curvelet transform, we obtain the reconstructed band image.}
    \label{fig:schem_curvelet}
\end{figure}

We applied a curvelet transform-based image sharpening algorithm to improve the quality of interpolated square images, which showed increased blurring along the y-axis with greater sampling distances. This method, inspired by our research on multi-modal fusion to enhance the spatial resolution of FTIR images\cite{mankar2021multi}, was adapted to increase O-PTIR imaging speed through sparse sampling along the y-axis. The algorithm effectively enhances images by incorporating spatial information from high spatial resolution band images into lower spatial resolution images, aligning the quality with that of high-resolution images. Our previous multi-modal image fusion study employed dark-field imaging to capture high-resolution spatial information, circumventing the diffraction-limited spatial resolution of FTIR imaging\cite{mankar2021multi}. Given that O-PTIR can achieve a resolution of \SI{0.5}{\micro\meter}, it allows us to avoid the previous challenges associated with integrating data from two distinct technologies, enabling the reconstruction of high-resolution \SI{0.5}{\micro\meter} $\times$ \SI{0.5}{\micro\meter} band images solely from sparse O-PTIR data. Furthermore, data from multiple O-PTIR bands are co-registered at acquisition. We initially perform linear equalization between the Amide I band image and each interpolated band image to preserve spectral information and adjust for absorption across different bands. Following equalization, we employ CurveLab 2.1.2 to reconstruct the interpolated image using the high-resolution image. We acquire the curvelet transform of the interpolated and Amide I images and combine the low-frequency components from the interpolated image while selecting high-frequency components from the Amide I image, resulting in sharper edges in the reconstructed image. We compute the inverse curvelet transform on the combined data to get the sharpened high-resolution band image. The schematic is presented in Figure \ref{fig:schem_curvelet}.

\begin{figure}[!hbt]
    \centering
  \includegraphics[width=\linewidth]{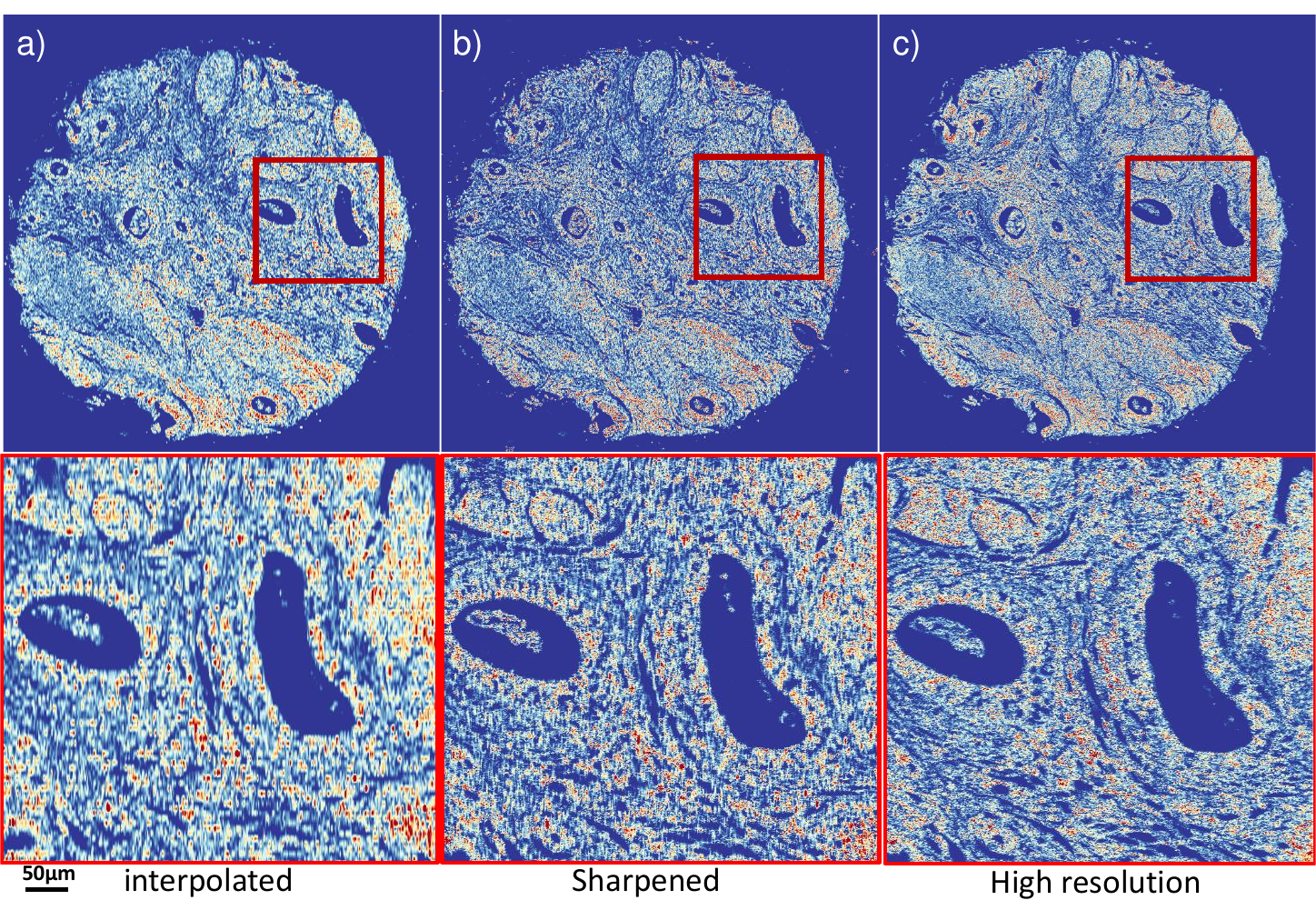}
    \caption{
    Comparison of (a) interpolated, (b) computationally reconstructed, and (c) experimentally obtained high-resolution images. The comparison reveals that data collected at higher speeds with lower resolution can be effectively compensated for by our image sharpening method, which significantly improves upon the interpolated image.}
    \label{fig:Recon-qual}
\end{figure}

The sharpened image exhibits superior edge delineation compared to the interpolated image, as demonstrated in Figure \ref{fig:Recon-qual}. The high-resolution image, experimentally collected at a pixel size of \SI{0.5}{\micro\meter} $\times$ \SI{0.5}{\micro\meter}, shows edges and intensities akin to those in the reconstructed image. In contrast, the interpolated image appears blurred, with smoother edges, potentially diminishing the accuracy of CNN networks that rely on both spectral and spatial information.

\subsection{Data annotation}
Based on H\&E-stained microscopy data, two pathologists independently classified tissue cores as \textit{stroma}, \textit{epithelium}, or \textit{necrosis}. H\&E and IR images were manually aligned to generate annotated data for machine learning, and labels were subsequently transferred to O-PTIR images. The tissue microarray (TMA) was divided into two halves, ensuring an equal number of cores in each cohort: the right half was designated for training, while the left half was reserved for testing.

\subsection{Classification Models and Hyperparameters}
The hyperparameters for the random forest classifier and the convolutional neural network (CNN) remain consistent with those reported in our previous work \cite{gajjela2023leveraging}. The primary enhancements in this study involve expanding the input from five to twenty-seven bands and increasing the quantity of training and testing data. Details on the total number of pixels allocated for testing and training are provided in Table~\ref{tbl:pixelct}. 

\begin{table}[!hbt]
\centering
 \caption{
 Number of O-PTIR pixels in training and testing datasets separated by class. To create the training and testing cohorts, the TMA is divided in half. First, a small, random data set is chosen, and a classifier is optimized. To prevent class bias in training, equal numbers of pixels are selected from each class. $10,000$ O-PTIR pixels per class are used in the RF classifier and $400,000$ pixels per class for CNNs.}
    \label{tbl:pixelct}
  \begin{tabular}[htbp]{@{}lll@{}}
    \hline
    Class & Training & Testing\\
    \hline
    Epithelium & $11,242,103$ & $24,056,862$\\
    Stroma & $10,039,196$ & $17,565,714$\\
    Necrosis & $2,286,794$ & $583,072$\\
    \hline\hline
    Total & $23,568,093$ & $42,205,648$\\
    \hline
  \end{tabular}
\end{table}

\subsection{Implementation}

All data pre-processing, processing, training and testing were performed in Python using open-source software packages. The CNNs were implemented in Python with the Keras library~\cite{chollet2015keras}, and the random forest was implemented using the Scikit-learn library.~\cite{scikit-learn}. An GeForce RTX 3090 GPU was used to measure the performance of the CNN classifier on five different sets of randomly selected training pixels.

%%%%%%%%%%%%%%%%%%
\section{Results}
\begin{figure}
    \centering
  \includegraphics[width=\linewidth,keepaspectratio]{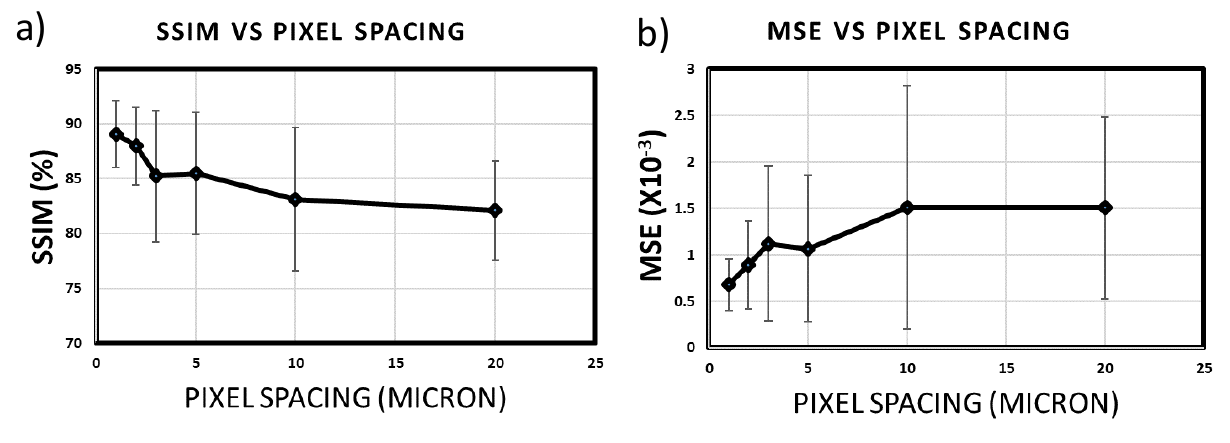}
    \caption{Reconstruction accuracy vs pixel spacing. 
    (a) Mean square error and (b) structural similarity (SSIM) averages vs pixel spacing. Data were collected from four cores at varying spacings along the Y-axis (\SI{0.5}{\micro\meter}×\SI{0.5}{\micro\meter}, \SI{0.5}{\micro\meter}×\SI{1}{\micro\meter}, \SI{0.5}{\micro\meter}×\SI{2}{\micro\meter}, \SI{0.5}{\micro\meter}×\SI{3}{\micro\meter}, \SI{0.5}{\micro\meter}×\SI{5}{\micro\meter}, \SI{0.5}{\micro\meter}×\SI{10}{\micro\meter}, \SI{0.5}{\micro\meter}×\SI{20}{\micro\meter}). The \SI{0.5}{\micro\meter}×\SI{0.5}{\micro\meter} spacing image served as a reference. We calculated MSE and SSIM for the various spacings and reported the mean and standard deviation for the cores.  }
    \label{fig:MSE}
\end{figure}

%%%%%%%%%%%%%%%%%%
We calculated the mean square error (MSE) and structural similarity index (SSIM) across various pixel spacings for four cores. SSIM evaluates the spatial feature similarity between reconstructed and original data, aiming for values near 1 for high similarity. MSE measures the average pixel error, with lower values indicating better reconstruction. The means and standard deviations of these metrics are depicted in Figure \ref{fig:MSE}. Both plots indicate that a pixel spacing of \SI{0.5}{\micro\meter} by \SI{5}{\micro\meter} yields favorable results compared to larger pixel spacings. While smaller pixel spacings lead to improved outcomes, a balance must be struck between data collection efficiency and reconstruction accuracy. Therefore, we recommend a pixel spacing of \SI{0.5}{\micro\meter} by \SI{5}{\micro\meter} as an optimal parameter for data collection using this technique.

Overall accuracy (OA) and receiver operating characteristic (ROC) curves were used to evaluate classifier performance. OA represents the percentage of pixels mapped correctly to the appropriate class for binary and multi-class classification. A ROC curve illustrates the correlation between specificity and sensitivity for identifying acceptable false positives and true positives. 

\begin{figure}[hbt]
    \centering
  \includegraphics[width=\linewidth]{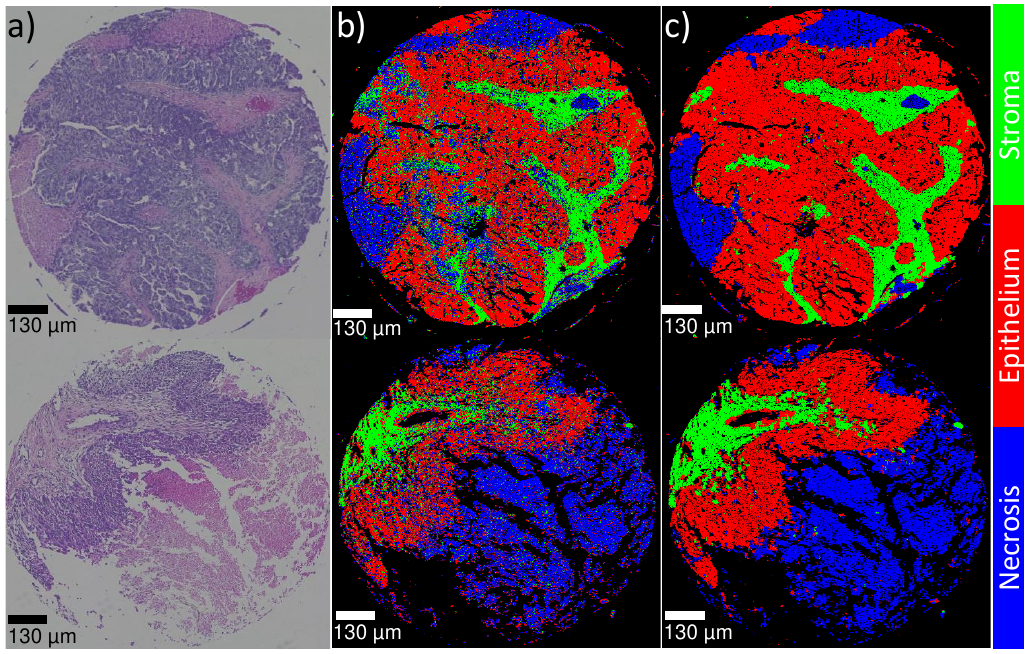}
    \caption{Comparison of stained image (a) identified as ground truth by a pathologist, with classifications by RF (b) and CNN (c) for two ovarian tissue cores. The RF model demonstrates significant improvement over our previously published results, attributed to the increased number of bands. Conversely, the CNN model achieves classification comparable to that of a pathologist's analysis on a stained tissue microArray (TMA), owing to its utilization of both spectral and spatial information. }
    \label{fig:Cores}
\end{figure}

We performed tissue segmentation using the Random Forest (RF) classifier, which leverages spectral information, and Convolutional Neural Networks (CNN), which utilize both structural and spectral information. The overall and class-wise accuracies for the testing dataset are detailed in Table ~\ref{tbl:rf-cnn-accuracy}. Compared to our previous work \cite{gajjela2023leveraging}, the accuracy of the RF classifier improved by approximately 35\%, attributed to the increase in the number of band images from five to twenty-seven. This expansion provides the RF classifier with more spectral information, leading to higher accuracies. As anticipated, CNNs surpass RF in performance, benefiting from their ability to incorporate structural information.

\begin{table}[hbt!]
\centering
 \caption{
Accuracy scores for the classification of Epithelium, Stroma, and Necrosis using (a) Random Forest (RF) and (b) Convolutional Neural Networks (CNNs) were averaged across five repetitions. The superior classification accuracy of CNNs can be attributed to their ability to leverage both spatial and spectral features, thereby outperforming RFs, which rely solely on spectral features. We determined the overall accuracy by calculating the weighted average accuracy of the classes.
 }
    \label{tbl:rf-cnn-accuracy}
  \begin{tabular}[htbp]{@{}llll@{}}
    \hline
    Class  & RF & CNN \\
    \hline
    Epithelium  & $82.3 \pm 0.2$ & $97.33 \pm 1.52$\\
    Stroma  & $73.8 \pm 0.2$ & $94.00 \pm 1.97$\\
    Necrosis  & $72.6 \pm 0.1$ & $83.00 \pm 1.97$\\
    \hline\hline
    Total  & $87.63 \pm 0.2$ & $95.735 \pm 0.82$\\
    \hline
  \end{tabular}
\end{table}

\begin{figure}[!hbt]
    \centering
  \includegraphics[width=\textwidth,keepaspectratio]{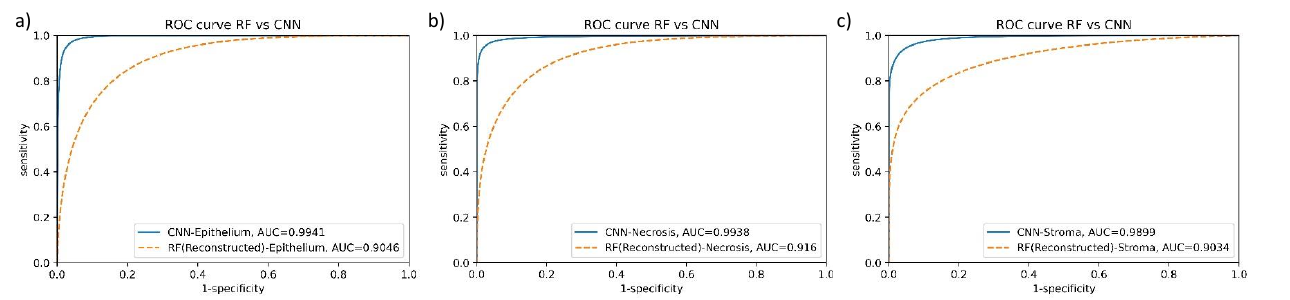}
    \caption{ROC curves and associated AUC values for each tissue subtype. CNN (blue line) demonstrates superior results compared to RF (dashed orange line) across all tissue subtypes: (a) epithelium, (b) necrosis, and (c) stroma.}
    \label{fig:ROC}
\end{figure}

Results that characterize the performance of all classifiers, as demonstrated by the Area Under the Curve (AUC) in a Receiver Operating Characteristic (ROC) plot, are presented in Figure \ref{fig:ROC}. Note that CNNs outperform RFs across all classes. This superiority of CNNs, attributed to their utilization of spatial features, which RFs lack, underscores the significance of integrating spatial and spectroscopic information to enhance tissue classification accuracy.

\section{Discussion}
O-PTIR technology enhances spectral data resolution from  $\approx5 \mu m$ to $0.5 \mu m$, outperforming current state-of-the-art FTIR systems. This advancement results in a 100$\times$ increase in pixel count over the same sample area, offering unprecedented spatial features and chemical information beyond what existing IR spectroscopic techniques can provide. However, this high resolution comes at the cost of slower data acquisition speeds, limited by the signal-to-noise ratio (SNR) and the stage speed of commercial O-PTIR imaging systems. Therefore, optimizing the hyperspectral data collection process for O-PTIR is essential.

To address this challenge, we implemented sparse, interleaved sampling along the Y-axis while maintaining the sampling rate at the diffraction limit in the X-direction. Although various robust sampling methods, such as random and Lissajous sampling, are viable for data reconstruction, the commercial system's design facilitates rapid acquisition along the X-axis at high pixel density, but acquisition is slow along the Y-axis. Given these constraints, we opted for interleaved Y-sampling as the most efficient strategy to collect sparse data.

\begin{figure}
    \centering
  \includegraphics[width=\textwidth,keepaspectratio]{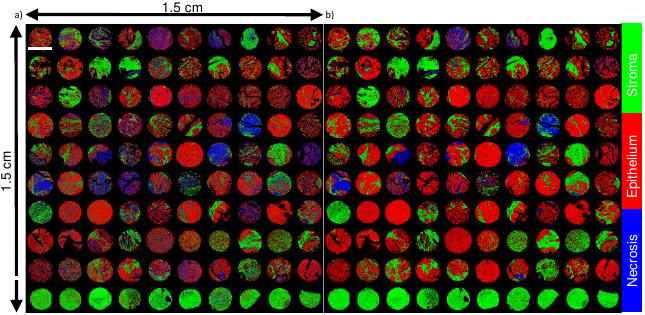}
    \caption{Classification results of 100 cores with (a) RF and (b) CNN. Red, Green, and Blue channels correspond to epithelium, stroma, and necrosis respectively. The scale bar is 1.5 mm }
    \label{fig:Full result}
\end{figure}

The size of pixels chosen for sampling relies on striking a balance between the time required to obtain data and the accuracy of data reconstruction. Table \ref{tab: X-Y sampling vs. imaging time} outlines the time it takes to collect data for a specific area, with sampling pixel sizes ranging from 0.5 to 10 $\mu m$. In these sets of samples, collecting each band image at full resolution would take approximately 37 hours; therefore, for 28 band images for each core, it would take approximately 37 hours at $0.5\times0.5 \mu m$ pixel spacing. In our method, by collecting 27 bands at $0.5\times5 \mu m$, and one band at maximum resolution for reconstruction purposes, each core image takes about 5 hours to collect. This shows that data collection alone is shortened almost 7 times, including all overheads. To determine the best pixel spacing, we used three key metrics, namely SSIM, MSE, and classification accuracy, to compare reconstructed data with high-resolution data from the O-PTIR system. These metrics were evaluated on five cancer cores selected randomly from an ovarian TMA. We limited the evaluation to only five cores because acquiring the 27 high-resolution images for each core can take anywhere from 24 to 36 hours, depending on core size. Therefore, obtaining high-resolution images for all 100 cores to compute these metrics is impractical. The SSIM metric measures the similarity in spatial features between the reconstructed data and the original raw data, aiming for a value close to 1 to indicate high similarity. The MSE measures the mean pixel-wise discrepancy between the reconstructed and original images, with values approaching zero indicating better reconstruction quality. To verify the accuracy of the reconstructed data in representing biological features, we employed random forest and Convolutional Neural Networks (CNNs) to determine whether these supervised machine learning algorithms could effectively distinguish between different cell types within ovarian and cervical tissues.

In our previous study, we classified epithelium and stroma in ovarian tissue using images from five specific wavenumbers with both random forest and CNN algorithms \cite{gajjela2023leveraging}. However, the 5-bands have insufficient spectroscopic information for identifying classes beyond epithelium and stroma, and the need for a broader range of wavenumbers became apparent. Therefore, we needed a broader range of wavenumbers. Here, we present a generalized approach to practically obtain hyperspectral data that opens new possibilities using O-PTIR. Inspired by results \cite{mankar2018selecting} that indicate that a whole (1600 band) hyperspectral data cube is unnecessary for multi-class classification in FT-IR, we selected 27 wavenumbers to achieve efficient multi-class classification. 

Our study reveals that a CNN significantly outperformed a random forest classifier, primarily because the latter depends on pixel-wise spectral data, whereas CNNs leverage spatial in addition to spectral features. The addition of 22 new reconstructed band images significantly improved the classification performance of the random forest classifier as demonstrated in Figure \ref{fig:ROC} and Table \ref{tbl:rf-cnn-accuracy}. The performance of random forest demonstrated the need for more spectral information, but obtaining additional band images would significantly increase the data acquisition time. 

Comparing our results to previous research \cite{gajjela2023leveraging} on the binary classification of ovarian cancer, we observed that the accuracy of the random forest classifier increased from 53\% with 5 bands to 87\% with 27 bands, thanks to the richer spectral information. Similarly, using CNNs led to an accuracy improvement from 90\% to 95\% when employing 27 bands. Note that our results from multi-class classification outperforms prior binary classification, which is a testament to the robustness and effectiveness of our approach. The classification outcomes for each algorithm are depicted in Figure \ref{fig:Full result}, with red, green, and blue channels representing epithelium, stroma, and necrosis, respectively. Additionally, Figure \ref{fig:Cores} shows two cores containing all classes. An adjacent section, stained with H\&E and annotated by a pathologist, shows a close alignment with our classification results.

\section{Conclusion}
We propose a novel high-speed Mid-Infrared Spectral Imaging (MIRSI) approach that reconstructs hyperspectral images using curvelets, addressing the significant bottleneck of data acquisition time in traditional MIRSI imaging. This technique involves acquiring sparse, interleaved data and applying reconstruction algorithms to overcome the challenges associated with slow data acquisition rates. By selecting higher-order curvelet coefficients from the Amide I image, our algorithm effectively reconstructs missing spatial information in sparse hyperspectral data, resulting in sharper edges and enhanced delineation of tissue features. To validate our approach, we employed several metrics, including MSE, SSIM, and tissue classification, to evaluate our method's capability in categorizing different cell types within an ovarian biopsy. Our technique improves O-PTIR data acquisition speed by 10X, making label-free histopathology practical. We have validated our approach extensively on statistically robust datasets with 100 ovarian cancer patients and >65 million data points. This work is a crucial step towards quantitative, label-free, automated histopathology and will be an invaluable tool for early cancer detection and comprehensive evaluation of ovarian tissue.

\medskip
%\textbf{Supporting Information} \par %Please delete the Suppporting Information statement if it is not applicable. Please supply Supporting Information in another file. Supporting information should not be provided in .tex format
%Supporting Information is available from the Wiley Online Library or from the author.

% Acknowledgements
%\medskip
%\textbf{Acknowledgements} \par %delete if not applicable))

\section*{Author Contributions}

\textbf{RRS}: Data curation, Visualization, Formal analysis, Investigation, Methodology, Software. \textbf{CCG}: Data curation, Visualization, Investigation, Writing- Original draft preparation. \textbf{XW}: Methodology, Investigation, Visualization. \textbf{RI}: Resources, Software. \textbf{SC}: Data curation, writing. \textbf{JL} \textbf{YG}: Data curation, \textbf{JL}: Resources, Data Curation. \textbf{AKS}: Conceptualization, Supervision. \textbf{DM}: Supervision, Writing- Reviewing and Editing. \textbf{SB}: Investigation, Methodology, Software, Validation. \textbf{RR}: Conceptualization, Funding acquisition, Project administration, Resources, Supervision, Writing- Reviewing and Editing.

\section*{Conflicts of interest}
There are no conflicts to declare.

\section*{Acknowledgements}
This work is supported in part by the Cancer Prevention and Research Institute of Texas (CPRIT) \#RR170075 (RR), NLM Training Program in Biomedical Informatics and Data Science \#T15LM007093 (RR and SB), National Institutes of Health \#R01HL146745 (DM), the National Science Foundation CAREER Award \#1943455 (DM).

%%%END OF MAIN TEXT%%%


\balance


%%%REFERENCES%%%
\bibliography{refs} %You need to replace "rsc" on this line with the name of your .bib file

\end{document}